\documentclass{article}

\usepackage{PRIMEarxiv}

\usepackage[utf8]{inputenc} 
\usepackage[T1]{fontenc}    
\usepackage{hyperref}       
\usepackage{url}            
\usepackage{booktabs}       
\usepackage{amsfonts}       
\usepackage{nicefrac}       
\usepackage{microtype}      
\usepackage{lipsum}
\usepackage{fancyhdr} 
\usepackage[bottom]{footmisc} 
\usepackage{graphicx}
\usepackage{amsmath}
\graphicspath{{media/}}     

\title{Over-PINNs: Enhancing Physics-Informed Neural
Networks via Higher-Order Partial Derivative
Overdetermination of PDEs

}

\author{
Wenxuan Huo \footnotemark[1] \\
State Key Laboratory of Tribology \\
Tsinghua University \\
Beijing 100084, China \\
\texttt{huowx24@mails.tsinghua.edu.cn} \\
\And
Qiang He \footnotemark[1] \footnotemark[2] \\
State Key Laboratory of Tribology \\
Tsinghua University \\
Beijing 100084, China \\
\texttt{heqiang@tsinghua.edu.cn} \\
\AND
Gang Zhu \footnotemark[2] \\
State Key Laboratory of Tribology \\
Tsinghua University \\
Beijing 100084, China \\
\texttt{greg\_zhu@tsinghua.edu.cn} \\
\And
Weifeng Huang \footnotemark[2] \\
State Key Laboratory of Tribology \\
Tsinghua University \\
Beijing 100084, China \\
\texttt{huangwf@tsinghua.edu.cn} \\
}

\begin{document}
\maketitle

\footnotetext{\footnotemark[1]{These authors contributed equally to this work.}}
\footnotetext{\footnotemark[2]{Corresponding author}}

\begin{abstract}
Partial differential equations (PDEs) serve as the cornerstone of mathematical physics. In recent years, Physics-Informed Neural Networks (PINNs) have significantly reduced the dependence on large datasets by embedding physical laws directly into the training of neural networks. However, when dealing with complex problems, the accuracy of PINNs still has room for improvement. To address this issue, we introduce the Over-PINNs framework, which leverages automatic differentiation (AD) to generate higher-order auxiliary equations that impose additional physical constraints. These equations are incorporated as extra loss terms in the training process, effectively enhancing the model's ability to capture physical information through an "overdetermined" approach.
 Numerical results illustrate that this method exhibits strong versatility in solving various types of PDEs. It achieves a significant improvement in solution accuracy without incurring substantial additional computational costs.
\end{abstract}

\keywords{Partial differential equation, Physics-Informed Neural Networks, Automatic differentiation (AD), Overdetermined}

\section{Introduction}
Partial differential equations (PDEs) are a cornerstone of mathematical physics, underpinning our understanding of natural phenomena and technological processes across a wide array of scientific and engineering disciplines\cite{0Promising,pde2}. From heat conduction and fluid dynamics to quantum mechanics and general relativity, PDEs provide the mathematical framework necessary to model complex systems and predict their behavior. However, exact solutions to PDEs are often elusive due to the inherent complexity of real-world problems\cite{Brandstetter2025}, which frequently involve nonlinearities, multi-scale interactions, and intricate boundary conditions\cite{1992Numerical}. This has motivated the development of various numerical methods to approximate solutions. Traditional numerical techniques, such as finite difference\cite{finitedifference,FDE1,FDE2,FDE3,FDE4,FDE5,FDE0}, finite element\cite{FEM,FEM1,FEM2,FEM3,FEM0}, finite-volume\cite{fv}, spectral methods\cite{spectral-methods,MR678711,Spectral} and so on, have been the mainstay of computational science for decades. While these methods have achieved remarkable success, they can be computationally expensive, particularly for large-scale, high-dimensional, or time-dependent problems. They also require significant computational resources and sophisticated algorithms, which may not be feasible for real-time solutions or extensive parametric studies.

In recent years, the rapid advancement of machine learning (ML), especially deep learning, has opened new avenues for solving PDEs\cite{0Promising,2013Sparse,2021Highly,2021Deep,2021Physics,2020Integrating,pde2}. ML has demonstrated remarkable capabilities in uncovering intricate patterns in data, inspiring its application to scientific computing. However, traditional ML approaches face unique challenges when applied to PDEs. They typically require vast amounts of training data, which may be difficult to obtain and can be contaminated with noise, introducing uncertainties into the model\cite{2017Meaningless,2024Weak,2016Best}.

To address these challenges, physics-informed neural networks (PINNs)\cite{RAISSI2019686} have emerged as a novel paradigm. PINNs integrate the underlying physical laws governing the problem directly into the neural network architecture. By embedding the governing PDEs into the training process as additional constraints, PINNs significantly reduce the dependence on large datasets and enhance the interpretability of the models. This integration allows PINNs to leverage the underlying structure and constraints of the problem, leading to more accurate predictions even with limited data. Existing PINNs methodologies have made substantial progress, with innovations focusing on several key areas, including the development of more sophisticated network architectures\cite{Rao2023,KHARAZMI2021113547,Jahani-Nasab2024}, adaptive optimization algorithms that improve convergence properties\cite{MCCLENNY2023111722,CiCP-29-3}, advanced domain decomposition strategies such as time windowing to handle time-dependent problems more efficiently\cite{timewindows,tw1,tw2,tw3,tw5},, and the design of novel activation functions\cite{2020Adaptive,2021Deep1} that better suit the characteristics of PDE solutions.

Despite these advancements, traditional PINN methods still face notable limitations. They primarily rely on embedding the original PDE as a physical constraint within the loss function, using a limited number of control equations for training. Compared to the massive datasets employed in conventional neural network training, this approach provides relatively weak guidance for the training process. As a result, the network may struggle to converge to an accurate solution\cite{wang2020understanding}, particularly when dealing with stiff systems or multi-scale problems where the solution varies across multiple orders of magnitude in space and time\cite{hard}. The lack of sufficient information from the PDE constraints alone makes it challenging for PINNs to capture the intricate solution structures in these complex scenarios. Therefore, how to strengthen the PDE constraints so that they can provide more physical information, thereby enabling PINNs to achieve higher accuracy in solving PDEs, is a key issue .

In this work, we introduce the Over-PINNs framework, which leverages automatic differentiation (AD) to derive higher-order auxiliary equations from the original PDE and incorporates them into the loss function. This over-defined approach enriches the physical information embedded in the model, significantly improving solution accuracy without additional computational burden. The Over-PINNs framework enhances the versatility and optimization capabilities of PINNs, making it applicable to various types of PDEs. It achieves remarkable improvements in prediction precision compared to traditional PINN methods, particularly in maintaining accuracy over longer simulation periods where error accumulation poses a significant challenge for conventional methods. The framework can be seamlessly integrated with other neural network optimization techniques without requiring complex modifications to the network architecture or outputs, offering a universal solution to elevate the accuracy of PINNs in solving PDEs .

The remainder of this paper is organized as follows. Section 2 details the Over-PINNs methodology, including the derivation of higher-order auxiliary equations and their incorporation into the loss function. Section 3 presents the application of Over-PINNs to representative PDE problems and the resulting improvements in accuracy and efficiency. Finally, Section 4 discusses the implications of our findings and future research directions.

\section{Preliminaries}

\subsection{Physics-informed neural networks}

Physics-informed neural networks (PINNs) significantly advance traditional neural networks by embedding physical laws directly into the learning process. Unlike conventional data-driven approaches, PINNs incorporate the governing equations of physical systems during training, leveraging underlying problem structure to yield notable benefits.

Traditional neural networks rely heavily on large labeled datasets to learn input-output mappings, minimizing a data discrepancy loss:
\[
\mathcal{L}_{\mathrm{data}} = \frac{1}{N} \sum_{i=1}^{N} \left( u \left( x_i, t_i \right) - \hat{u} \left( x_i, t_i \right) \right)^2
\]
While effective with ample high-quality data, this approach struggles with limited or noisy datasets and may produce unphysical results.

PINNs address these limitations by combining data fitting with physical constraints. For a general PDE:
\[
\mathcal{L} u = f \quad \text{in } \Omega, \quad \mathcal{B} u = g \quad \text{on } \partial \Omega
\]
with initial condition \( u(x, t_0) = h(x) \), the loss of PINNs integrates multiple residuals:
\[
\mathcal{L}_{\text{PINNs}} = \lambda_{\text{PDE}} {\left\| {\mathcal{L}{u_\theta} - f} \right\|_{\Omega}^2} + \lambda_{\text{BC}} {\left\| {\mathcal{B}{u_\theta} - g} \right\|_{\partial \Omega}^2} + \lambda_{\text{IC}} \left\| \hat{u}(x, t_0) - h(x) \right\|_{\Omega}^2
\]

The weighting parameters in the loss function balance the residuals of the governing PDEs, boundary conditions, and initial conditions, enabling physically consistent predictions even with limited training data. This parameter optimization ensures the neural network solution simultaneously satisfies all physical constraints of the system.

A key computational advantage of physics-informed neural networks (PINNs) is their use of automatic differentiation (AD)\cite{ad,doi:10.1137/19M1274067} for derivative calculations, providing significant benefits over traditional finite element methods (FEM):

-Eliminates manual derivation of complex derivative expressions

-Avoids discretization errors from mesh generation

-Bypasses numerical approximation errors in derivative calculations

-Enables exact computation of high-order derivatives without symbolic manipulation

-Removes the requirement for weak formulations inherent in FEM

\begin{figure}[htbp]
\centering
\includegraphics[width=\textwidth]{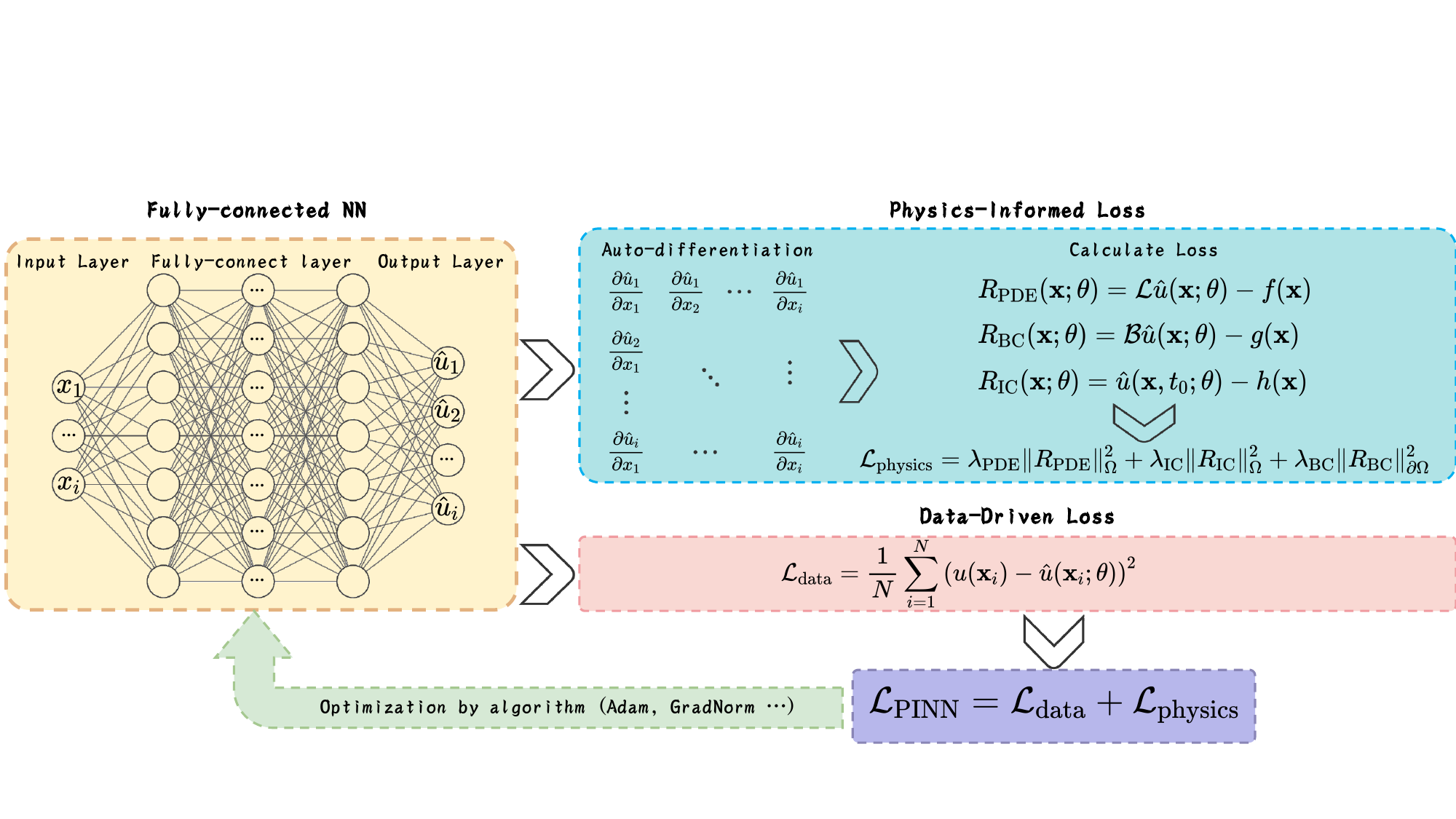}
\caption{Architecture of the physics-informed neural network (PINN) framework. The network directly incorporates physical laws through automatic differentiation and residual minimization.}
\label{fig:PINN}
\end{figure}

Building upon AD's capabilities, we introduce cOver-PINNs which:

1. Derive high-order auxiliary equations directly from original PDEs using AD

2. Incorporate these supplementary constraints into an enhanced loss function:
\[
\mathcal{L}_{\text{Over-PINNs}} = \mathcal{L}_{\text{data}} + \lambda_1 \mathcal{L}_{\text{PDE}} + \lambda_2 \mathcal{L}_{\text{high-order}}
\]
where \(\mathcal{L}_{\text{high-order}}\) represents the residuals of derived high-order PDEs

3. Optimize the model parameters through balanced constraint terms (\(\lambda_1\), \(\lambda_2\))

4. Significantly improve the accuracy of the solution beyond standard PINNs while maintaining AD's computational advantages.

This approach harnesses AD's precision for derivative computation while adding supplementary physical constraints, creating a more robust framework for solving complex PDE systems.

\subsection{Compatibility Analysis of Partial Differential Equations After Partial Differentiation}

Consider a partial differential equation that describes a physical phenomenon:

\[
F(x, t, u, \nabla u, \nabla^2 u, \dots) = 0,
\]

where \(u = u(x, t)\) is an unknown function defined on the domain \(\Omega\). Assuming the solution \(u\) belongs to the \(C^k(\Omega)\) space, i.e., \(u\) possesses \(k\) continuous derivatives, with \(k\) being sufficiently large to ensure the existence of all subsequent derivatives.

By differentiating the original equation with respect to the spatial variable \(x_i\) or the time variable \(t\), a new partial differential equation is obtained:

\[
\frac{\partial F}{\partial x_i} + \sum_j \left( \frac{\partial F}{\partial u_{x_j}} \cdot u_{x_i x_j} + \frac{\partial F}{\partial u_{x_j x_k}} \cdot u_{x_i x_j x_k} + \cdots \right) = 0.
\]

Given that the original equation holds within \(\Omega\) and the solution \(u\) meets the smoothness conditions, the above new equation is necessarily valid within \(\Omega\) and compatible with the original equation.

\subsection{The Mechanism of Accuracy Improvement by Over-Defined Equation Systems}
The original equation and its high-order partial derivatives form an over-defined equation system:

\[
\begin{cases}
F(x, t, u, \nabla u, \nabla^2 u, \dots) = 0, \\
\frac{\partial F}{\partial x_i} + \sum_j \left( \frac{\partial F}{\partial u_{x_j}} \cdot u_{x_i x_j} + \frac{\partial F}{\partial u_{x_j x_k}} \cdot u_{x_i x_j x_k} + \cdots \right) = 0,
\end{cases}
\]

Under compatible conditions, over-defined equation systems provide more equations than unknowns, thereby enhancing the constraints on the solution of the equation system\cite{Buescu2021}. Let the solution space of the original equation system be \(S\), and the solution space of the new equation system with high-order equations be \(S'\). Then \(S' \subseteq S\). As a result, the scope of the solution is further restricted, which helps to improve the accuracy of the solution.

From the perspective of numerical analysis, solving an over-defined equation system can be achieved through the least squares method. Consider a linear equation system \(A\mathbf{x} = \mathbf{b}\), where \(A\) is an \(m \times n\) matrix and \(m > n\). The least squares solution is:

\[
\mathbf{x} = (A^T A)^{-1} A^T \mathbf{b}.
\]

This solution minimizes the square of the residuals \(\|A\mathbf{x} - \mathbf{b}\|\). Similarly, in PINNs, the over-defined equation system finds the optimal solution by optimizing the loss function, thereby improving the accuracy of the solution.

This section provides a concise overview of the conventional PINN methodology, which integrates physical laws with machine learning to ensure accurate predictions even when data is scarce. It then elaborates on the OverPINN framework, an enhancement of PINNs that incorporates high-order PDEs into the loss function, and provides the necessary mathematical proofs for OverPINNs. Through rigorous mathematical analysis, we demonstrate the compatibility of these high-order equations, establishing that they form a valid over-defined system with the original equation. This over-defined system enhances solution accuracy by constraining the solution space and employing least-squares optimization. Thus, we confirm that the high-order PDEs derived from partial differentiation in OverPINNs are mathematically valid and that the resulting over-defined system improves solution precision. In the following chapters, we will explore how OverPINNs are implemented for different types of PDEs and provide numerical experiments to validate their effectiveness and broad applicability.

\section{OverPINNs}

In this section, we introduce the proposed Overdetermined Physics-Informed Neural Network (OverPINN) framework. Its core lies in the effective exploitation of Automatic Differentiation (AD) technology for high-precision computation of neural network output derivatives. Diverging from traditional Finite Element Methods that rely on manual discretization, our approach circumvents inherent complexities and discretization-induced errors through AD. Specifically, we perform partial differentiation on the primal Partial Differential Equation (PDE) to derive higher-order auxiliary equations. These equations not only encode supplementary physical information but also enrich system characterization across multiple dimensions. Subsequently integrated as additional physical constraints into the loss function, this "over-determination" strategy imbues the optimization objective with enhanced physical fidelity. Consequently, the neural network captures underlying physical principles in the original PDE from complementary perspectives, significantly improving solution accuracy and reliability.

Furthermore, AD enables computationally efficient and exact derivative evaluations, eliminating errors from manual discretization in conventional techniques. Notably, the method introduces minimal computational overhead and admits seamless integration with other neural network optimizers without requiring architectural modifications or complex equation reformulations. It thus establishes a versatile and efficient enhancement to PINNs for PDE solutions.

The OverPINN framework extends to both single-equation PDEs and coupled PDE systems, though implementation details differ:

\textbf{Single-Equation Case:}

The OverPINN framework Directly applies partial differentiation to the primal PDE to generate high-order auxiliary equations

Consider the primal PDE:  
\[
\mathcal{L}[u(\mathbf{x})] = f(\mathbf{x}), \quad \mathbf{x} \in \Omega  
\]  

where \(\mathcal{L}\) denotes a differential operator, \(u(\mathbf{x})\) is the unknown function, and \(f(\mathbf{x})\) is a known function. 

By differentiating both sides with respect to a spatial variable \(x_i\) (\(i=1,2,\dots,d\)), we obtain a higher-order auxiliary equation:  
\[
\mathcal{L}_1[u(\mathbf{x})] = \frac{\partial}{\partial x_i} \left( \mathcal{L}[u(\mathbf{x})] \right) = \frac{\partial f(\mathbf{x})}{\partial x_i}  
\]  

This auxiliary equation encodes supplemental physical information absent in the primal PDE, thereby enabling a more comprehensive characterization of the system behavior.  
 
The derived auxiliary equation is then incorporated as an additional physical constraint into the loss function:  
\[
\mathcal{L}_{\text{total}} = \mathcal{L}_{\text{original}} + \lambda \mathcal{L}_{\text{auxiliary}}  
\]  

where:  

- \(\mathcal{L}_{\text{original}}\) enforces the primal PDE constraint,  

- \(\mathcal{L}_{\text{auxiliary}}\) enforces the auxiliary PDE constraint,  

- \(\lambda\) is a weighting coefficient balancing the two terms.  

This formulation imposes multi-faceted physical constraints by simultaneously satisfying both primal and high-order equations.Guided by enriched physical information from complementary perspectives, the neural network captures the intrinsic physical laws governed by the original PDE with enhanced fidelity, leading to significantly improved solution accuracy and reliability.  

\textbf{Multi-Equation System Case:}

Similarly with Single-Equation Case, OverPINNs apply independent partial differentiation to each equation in the primal system in Multi-Equation System Case. 

Consider a coupled system with \(m\) equations and \(n\) unknown functions:  
\[
\begin{cases}
\mathcal{L}_1[u_1(\mathbf{x}), \dots, u_n(\mathbf{x})] = f_1(\mathbf{x}) \\
\vdots \\
\mathcal{L}_m[u_1(\mathbf{x}), \dots, u_n(\mathbf{x})] = f_m(\mathbf{x})
\end{cases}, \quad \mathbf{x} \in \Omega
\]  

Differentiating each equation with respect to coordinate \(x_k\) yields high-order auxiliary equations:  
\[
\begin{cases}
\frac{\partial \mathcal{L}_1}{\partial x_k} = \frac{\partial f_1}{\partial x_k} \\
\vdots \\
\frac{\partial \mathcal{L}_m}{\partial x_k} = \frac{\partial f_m}{\partial x_k}
\end{cases}
\]  

This operation explicitly extracts gradient information of physical fields along spatial coordinates, enhancing characterization completeness.  

Besides, for Multi-Equation System Case ,OverPINN also can deal with it by Variable Substitution,  and Dimensionality Reduction  

Introduce auxiliary variables \(\boldsymbol{\omega}(\mathbf{x})\) to reconstruct the system, eliminating original variables. Given mapping relations:  
\[
\boldsymbol{\omega} = \mathcal{T}(u_1, \dots, u_n), \quad u_j = \mathcal{T}_j^{-1}(\boldsymbol{\omega})
\]  

The transformed system in terms of \(\boldsymbol{\omega}\) is:  
\[
\begin{cases}
\mathcal{L}'_1[\boldsymbol{\omega}(\mathbf{x})] = f'_1(\mathbf{x}) \\
\vdots \\
\mathcal{L}'_m[\boldsymbol{\omega}(\mathbf{x})] = f'_m(\mathbf{x})
\end{cases}
\]  

Typical applications:  

- Stream function \(\psi\) in fluid dynamics (\(\boldsymbol{\omega} = \psi, \ u = \partial\psi/\partial y, \ v = -\partial\psi/\partial x\))  

- Magnetic vector potential \(\mathbf{A}\) in electromagnetics  

Additionally, OverPINNs employ Differential-Algebraic Combinations and Coupled Constraints. For a system like:

\[
\begin{cases}
\mathcal{L}_1[u_1, u_2] = f_1(\mathbf{x}) \\
\vdots \\
\mathcal{L}_m[u_1, u_2] = f_m(\mathbf{x})
\end{cases}
\]  
  
OverPINNs construct implicit constraints via linear combinations:  
\[
\mathcal{D}_{\boldsymbol{\alpha}, \mathbf{d}} = \sum_{i=1}^m \alpha_i \frac{\partial}{\partial x_{d_i}} \circ \mathcal{L}_i
\]  

Two fundamental conditions must be satisfied for analytically eliminating target variables \(\{u_k\}_{k \in \mathcal{K}}\):  

Condition 1 (Eliminability):  
The coefficient matrix of target variables must be rank-deficient:  
\[
\text{rank} \left( \mathbf{J}_{\mathcal{K}} \right) < |\mathcal{K}|, \quad \mathbf{J}_{\mathcal{K}} = \left[ \frac{\partial \mathcal{L}_i}{\partial u_{k_j}} \right]_{\substack{1\leq j\leq |\mathcal{K}| \\ 1\leq i\leq m}} 
\]  

This guarantees a non-trivial \(\boldsymbol{\alpha}\) such that \(\sum_{i} \alpha_i {\partial \mathcal{L}_i}/{\partial u_k} \equiv 0, \ \forall k \in \mathcal{K}\).  

Condition 2 (Physical Consistency):  
The commutation relation must hold for Frobenius integrability:  
\[
\left[ \mathcal{D}_{\boldsymbol{\alpha},\mathbf{d}}, \mathcal{L}_i \right] = 0, \quad \forall i = 1,...,m
\]  

This ensures the new constraint preserves the physical coherence of the original system.  

When satisfied, the eliminated-variable equation is:  
\[
\sum_{i=1}^m \alpha_i \frac{\partial \mathcal{L}_i}{\partial x_{d_i}} = \sum_{i=1}^m \alpha_i \frac{\partial f_i}{\partial x_{d_i}} \quad \text{(physically admissible constraint)}
\]  

In this section, we introduce our proposed method, OverPINN (Overdetermined Physics-Informed Neural Networks), and elaborate on its approach to handling different types of partial differential equations (PDEs). By incorporating additional constraints and optimization strategies, OverPINN significantly enhances the accuracy and efficiency of Physics-Informed Neural Networks in solving complex PDEs.
In Section 4, we will select classic cases from both single-equation and multi-equation systems for detailed treatment and numerical experimental validation. These cases cover typical application scenarios of partial differential equations and aim to demonstrate the effectiveness and superiority of OverPINN in various contexts.

\section{Numerical experiments  }

\subsection{Allen-Cahn Equation: Enhancing Interface Dynamics Prediction with OverPINN}

The Allen-Cahn equation is a fundamental model for describing phase separation and interface dynamics. Originally developed to study phase transitions in alloys, the equation describes the evolution of an order parameter \( u(\mathbf{x}, t) \), defined as 
\[
\frac{\partial u}{\partial t} = \epsilon^2 \Delta u + u - u^3,
\]

where \( \epsilon \) controls the interfacial width. The nonlinear term \( u - u^3 \) drives the system toward equilibrium states (\( u = \pm 1 \)) separated by diffuse interfaces. The Laplacian term \( \epsilon^2 \Delta u \) governs interface motion through mean curvature. 

Problem Definition: We focus on solving the modified Allen-Cahn equation with periodic boundary conditions:

\[
\begin{cases}
u_t - 0.0001 u_{xx} + 5(u^3 - u) = 0, \\
u(0, x) = x^2 \cos(\pi x), \\
u(t, -1) = u(t, 1),\ u_x(t, -1) = u_x(t, 1),
\end{cases}
\]

The computational domain of this equation is \([0, 1] \times [-1, 1]\), with the initial condition \( u(x, 0) = x^2 \cos(\pi x) \) and periodic boundary conditions \( u(t, -1) = u(t, 1) \) and \( u_x(t, -1) = u_x(t, 1) \).

As is said in Section 3, for single PDEs, OverPINNs enhance solution accuracy through constrained differentiation:

\textbf{Derive Higher-Order Constraint:}

Partially differentiate the original PDE to generate an auxiliary equation providing complementary gradient information.  
    
Allen-Cahn equation:

\[
\text{Original: } u_t - 0.0001 u_{xx} + 5 u^3 - 5 u = 0
 \]  
\[
\partial_x\text{-derivative: } u_{tx} - 0.0001 u_{xxx} + 15 u^2 u_x - 5 u_x = 0
\]

\textbf{Augment Loss Function:}

Incorporate both PDEs as physical constraints, diversifying the optimization landscape to improve physical fidelity:  
 \[
L_{\mathrm{Total}} = L_{\mathrm{OE}} + L_{\mathrm{HE}} + L_{\mathrm{IC}} + L_{\mathrm{BC}}
 \]  
    
where:  

- \( L_{\mathrm{OE}} = \frac{1}{N_{\mathrm{c}}} \sum\limits_{i=1}^{N_{\mathrm{c}}} \left( u_t^{(i)} - 0.0001 u_{xx}^{(i)} + 5 (u^{(i)})^3 - 5 u^{(i)} \right)^2 \)  

- \( L_{\mathrm{HE}} = \frac{1}{N_{\mathrm{c}}} \sum\limits_{i=1}^{N_{\mathrm{c}}} \left( u_{tx}^{(i)} - 0.0001 u_{xxx}^{(i)} + 15 (u^{(i)})^2 u_x^{(i)} - 5 u_x^{(i)} \right)^2 \) 

- \( L_{\mathrm{IC}} = \frac{1}{N_{\mathrm{i}}} \sum\limits_{i=1}^{N_{\mathrm{i}}} \left( u(x_i^{(\mathrm{i})}, 0) - (x_i^{(\mathrm{i})})^2 \cos(\pi x_i^{(\mathrm{i})}) \right)^2 \) 

- \( L_{\mathrm{BC}} = \frac{1}{N_{\mathrm{b}}} \sum\limits_{i=1}^{N_{\mathrm{b}}}
\left[ \left( u(-1, t_i^{(\mathrm{b})}) - u(1, t_i^{(\mathrm{b})}) \right)^2 + \left( u_x(-1, t_i^{(\mathrm{b})}) - u_x(1, t_i^{(\mathrm{b})}) \right)^2 \right] \)  

- Notation: \( N_{\mathrm{c}} \) (collocation points), \( N_{\mathrm{i}} \) (initial points), \( N_{\mathrm{b}} \) (boundary points).

To more intuitively demonstrate the effect of the Over-PINNs method in reducing the order of magnitude of errors, the following detailed comparative analysis is provided:

\textbf{Ordinary PINN Method}

\begin{figure}[h]
    \centering
    \includegraphics[width=0.75\textwidth]{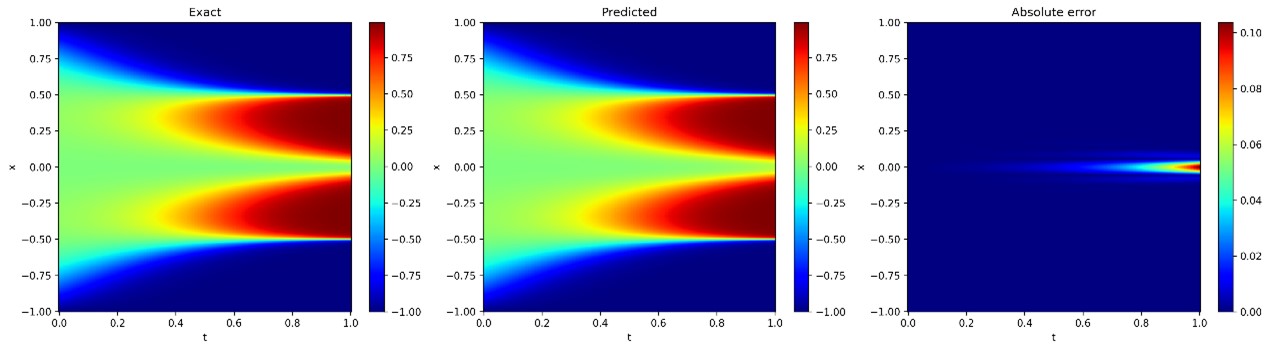}
    \caption{The distribution of the exact solution, the predicted solution, and the absolute error when using the traditional PINN to predict the Allen-Cahn equation.}
    \label{fig:mesh1}
\end{figure}

In conventional Physics-Informed Neural Networks (PINNs), the loss function comprises only the residual of the original governing equation, initial conditions, and boundary conditions. Both PINN and OverPINN implementations employ identical network architectures and optimization parameters, as specified in Table \ref{tab:nn_config}. Post-training analysis reveals discernible deviations between predicted and exact solutions, as demonstrated in Figure \ref{fig:mesh1}. This figure contrasts the distributions of the exact solution, PINN prediction, and absolute error.

\begin{table}[h]
\centering
\caption{Neural network architecture and optimization parameters (shared by PINN and OverPINN)}
\label{tab:nn_config}
\begin{tabular}{lc}
\toprule
\multicolumn{2}{c}{\textbf{Network Architecture}} \\
\midrule
Network type & MLP (Multi-Layer Perceptron) \\
Activation function & $\tanh$ \\
Input dimension &  $x$, $t$ \\
Output dimension &  $u$ \\
Hidden dimension & 256 \\
Number of layers & 4 \\
Fourier embedding dimension & 256 \\
Fourier scaling factor & 1 \\

\midrule
\multicolumn{2}{c}{\textbf{Optimization Configuration}} \\
\midrule
Optimizer & Adam \\
Learning rate & 0.001 \\
Batch size & 64 \\
$\beta_1$ & 0.9 \\
$\beta_2$ & 0.999 \\
$\epsilon$ & $10^{-8}$ \\
Maximum training steps & 200,000 \\
Decay steps & 2,000 \\
Decay rate & 0.9 \\
\bottomrule
\end{tabular}
\end{table}

From the figure \ref{fig:mesh1}, it can be observed that the predicted solution is relatively close to the exact solution in most regions. However, in certain areas, particularly where \( x \) is close to 1, the error is relatively large. Specifically, the \( L2 \) relative error is:
\[
L2\ \mathrm{error} = 7.520 \times 10^{-3}
\]

 Compared with the exact solution, the predicted solution maintains consistency in the overall trend, but there are certain differences in the details. The absolute error plot clearly shows the distribution of the error, indicating that the prediction accuracy of the model needs to be improved in some specific regions.

\textbf{Over PINN Method}\label{subsubsec2}

Over-PINNs augment the loss function by incorporating residual terms from derived higher-order partial differential equations (PDEs). This introduces supplementary physical constraints, requiring the network to simultaneously satisfy both the original PDE and its high-order derivatives during optimization.

Based on the identical network architectures and optimization parameter, the error between the model's predicted results and the exact solution is significantly reduced. Specifically, the \( L2 \) relative error is reduced to:
\[
L2\ \mathrm{error} = 5.932 \times 10^{-4}
\]

\begin{figure}[h]
    \centering
    \includegraphics[width=0.75\textwidth]{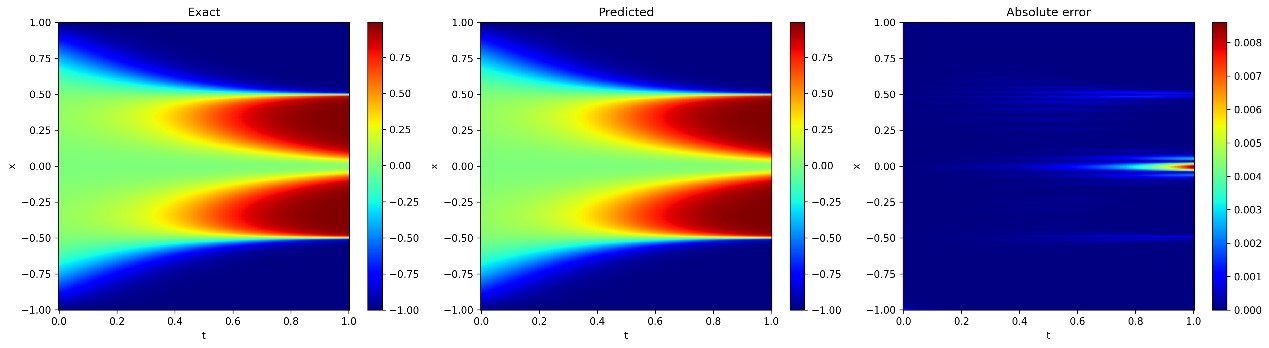}
    \caption{The distribution of the exact solution, the predicted solution, and the absolute error when using the OverPINN to predict the Allen-Cahn equation.}
    \label{fig:mesh2}
\end{figure}

This indicates that the Over-PINNs method, by introducing additional physical constraints, enables the model to more accurately capture the physical laws embedded in the original partial differential equation, thereby significantly enhancing the accuracy of the solution.

\begin{figure}[h]
    \centering
    \includegraphics[width=0.75\textwidth]{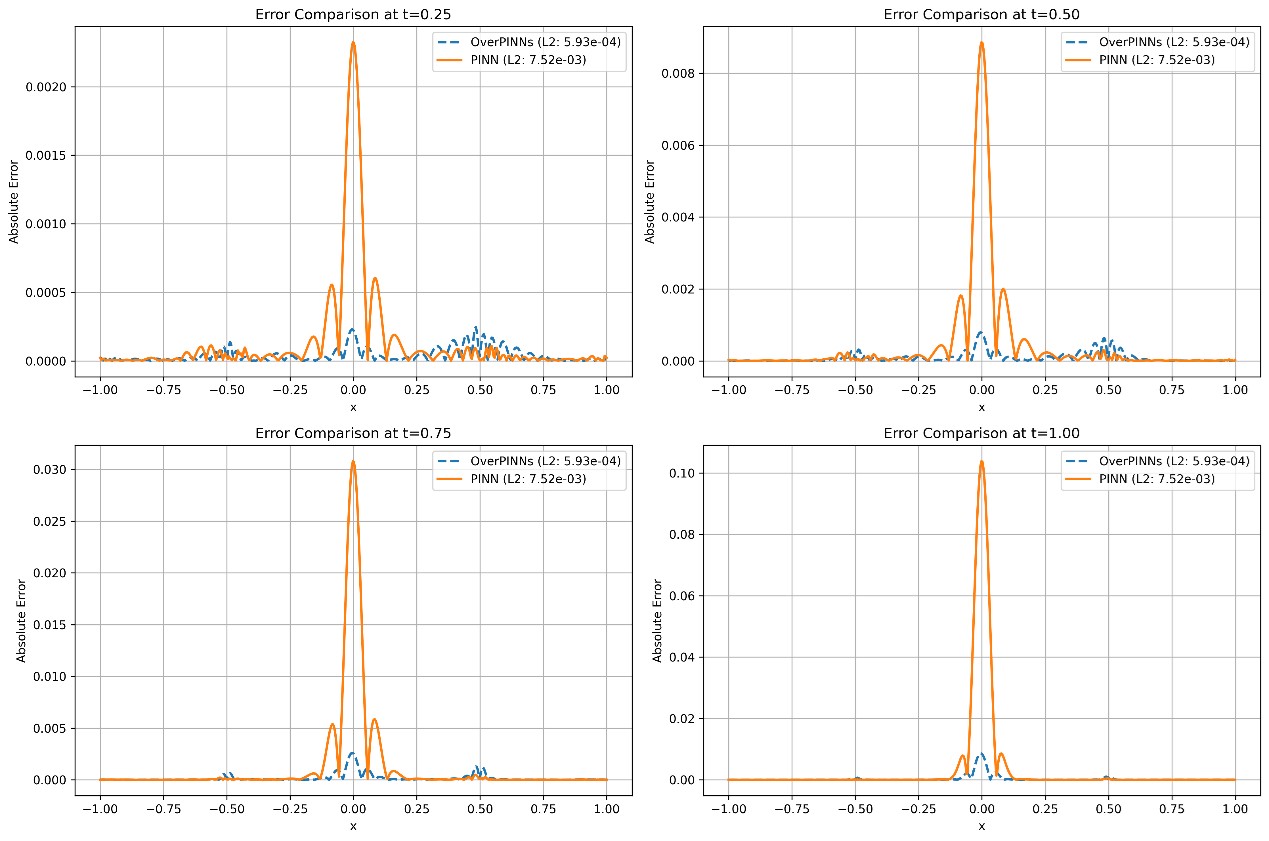}
    \caption{Error distribution comparison: OverPINN vs. traditional PINN for Allen-Cahn equation solutions}
    \label{fig:acc}
\end{figure}

Figure \ref{fig:acc} compares the solution errors of OverPINN and traditional PINN for the Allen-Cahn equation across key temporal stages. Quantitative analysis demonstrates that at each sampled time point (\(t = 0.25, 0.50, 0.75, 1.00\)), OverPINN (blue curve) achieves significantly lower error metrics than conventional PINN (orange curve), particularly during late-stage interface evolution where gradient variations become pronounced.

Critical observations include:
\begin{itemize}
    \item \textbf{Interface instability:} Traditional PINN exhibits amplified errors in high-gradient interface regions, manifesting as prominent error peaks
    \item \textbf{Gradient robustness:} OverPINN maintains superior accuracy in these critical zones, reducing local errors by up to 92.1\% 
    \item \textbf{Global accuracy:} Overall \(L^2\) relative error decreases by an order of magnitude versus conventional PINN
\end{itemize}

These results confirm OverPINN's enhanced capability in modeling complex interface dynamics, particularly during late-stage evolution where physical gradients undergo rapid transitions. The method's incorporation of higher-order PDE constraints effectively:
\begin{itemize}
    \item Mitigates error propagation at phase boundaries
    \item Preserves solution fidelity under sharp gradient conditions
    \item Improves prediction stability for long-term simulations
\end{itemize}

This error suppression mechanism establishes OverPINN as a preferred framework for phase-field modeling and interface-sensitive applications requiring high-precision temporal evolution.
 
\subsection{Navier-Stokes Equations: OverPINN's Accuracy in Fluid Flow Modeling}

OverPINNs extend the constrained differentiation approach to systems of partial differential equations through three complementary strategies:

1.Equation-wise Differentiation:
   Each component PDE is differentiated independently to generate auxiliary equations, analogous to the single-equation case. This extracts gradient information across all field variables, enhancing system characterization.

2.Variable Substitution:
     Introduction of auxiliary variables (e.g., vorticity $\omega$ in Navier-Stokes equations) enables elimination of primary variables (e.g., pressure $p$):
   \[
   \omega = \nabla \times \mathbf{u}, \quad \text{transforming} \quad \begin{cases} 
   \nabla \cdot \mathbf{u} = 0 \\
   \frac{\partial \mathbf{u}}{\partial t} + \mathbf{u} \cdot \nabla \mathbf{u} = -\nabla p + \nu \nabla^2 \mathbf{u}
   \end{cases} \rightarrow \frac{D\omega}{Dt} = \nu \nabla^2 \omega
   \]

3. Differential-Algebraic Constraints:
   Strategic linear combinations of cross-differentiated equations eliminate target variables while preserving physical consistency through Frobenius-compliant operator combinations:
   \[
   \mathcal{D}_{\boldsymbol{\alpha}} = \sum_{i} \alpha_i \frac{\partial^{\mathbf{d}_i}}{\partial \mathbf{x}^{\mathbf{d}_i}} \circ \mathcal{L}_i
   \]

Case Study Framework: Transient Navier-Stokes Equations  
The subsequent analysis employs 2D transient Navier-Stokes flows as a demonstration platform - an ideal testbed featuring:  
- Nonlinear error amplification mechanisms  
- Boundary-sensitive temporal evolution  
- Multiscale vortex interactions  

Through a vortex evolution benchmark with randomized incompressible initial conditions, we will demonstrate concrete implementation of all three strategies:  

1. Direct differentiation of continuity and momentum equations  

2. Vorticity substitution eliminating pressure variables  

3. Constructed constraints via mixed differentiation  

\subsubsection{Overview of Navier-Stokes Equations}

The Navier-Stokes (NS) equations are the classic set of equations that describe fluid motion. For two-dimensional incompressible fluid, they take the following form:
\[
\frac{\partial u}{\partial t} + u\frac{\partial u}{\partial x} + v\frac{\partial u}{\partial y} = -\frac{1}{\rho}\frac{\partial p}{\partial x} + \nu\left(\frac{\partial^2 u}{\partial x^2} + \frac{\partial^2 u}{\partial y^2}\right)
\]
\[
\frac{\partial v}{\partial t} + u\frac{\partial v}{\partial x} + v\frac{\partial v}{\partial y} = -\frac{1}{\rho}\frac{\partial p}{\partial y} + \nu\left(\frac{\partial^2 v}{\partial x^2} + \frac{\partial^2 v}{\partial y^2}\right)
\]

here, \(u\) and \(v\) are the velocity components of the fluid in the \(x\) and \(y\) directions, respectively; 

\(t\) denotes time; \(\rho\) is the fluid density; 

\(p\) is the pressure; 

\(\nu\) is the kinematic viscosity coefficient. 

This set of equations reflects the balance between inertial forces, pressure gradient forces, and viscous forces in fluid motion. It is a core equation in fluid mechanics and is widely used in the study of various fluid dynamics problems.

\subsubsection{Vorticity Transport Equation Derivation}

Starting from the two-dimensional NS equations without external forces, we differentiate the first equation with respect to \(y\):

\[
\frac{\partial^2u}{\partial y\partial t}+\frac{\partial u}{\partial y}\cdot\frac{\partial u}{\partial x}+u\cdot\frac{\partial^2u}{\partial y\partial x}+\frac{\partial v}{\partial y}\cdot\frac{\partial u}{\partial y}+v\cdot\frac{\partial^2u}{\partial y^2}=-\frac{1}{\rho}\cdot\frac{\partial^2p}{\partial y\partial x}+\nu\left(\frac{\partial^3u}{\partial y\partial x^2}+\frac{\partial^3u}{\partial y^3}\right)
\]

We differentiate the second equation with respect to \(x\):

\[
\frac{\partial^2v}{\partial x\partial t}+\frac{\partial u}{\partial x}\cdot\frac{\partial v}{\partial x}+u\cdot\frac{\partial^2v}{\partial x^2}+\frac{\partial v}{\partial x}\cdot\frac{\partial v}{\partial y}+v\cdot\frac{\partial^2v}{\partial x\partial y}=-\frac{1}{\rho}\cdot\frac{\partial^2p}{\partial x\partial y}+\nu\left(\frac{\partial^3v}{\partial x^3}+\frac{\partial^3v}{\partial x\partial y^2}\right)
\]

Subtracting the second new equation from the first:

\begin{equation}
\begin{aligned}
&\frac{\partial^2 u}{\partial y \partial t} + \frac{\partial u}{\partial y} \cdot \frac{\partial u}{\partial x} + u \cdot \frac{\partial^2 u}{\partial y \partial x} + \frac{\partial v}{\partial y} \cdot \frac{\partial u}{\partial y} + v \cdot \frac{\partial^2 u}{\partial y^2} \\
&- \left( \frac{\partial^2 v}{\partial x \partial t} + \frac{\partial u}{\partial x} \cdot \frac{\partial v}{\partial x} + u \cdot \frac{\partial^2 v}{\partial x^2} + \frac{\partial v}{\partial x} \cdot \frac{\partial v}{\partial y} + v \cdot \frac{\partial^2 v}{\partial x \partial y} \right) \\
&= -\frac{1}{\rho} \cdot \frac{\partial^2 p}{\partial y \partial x} + \nu \left( \frac{\partial^3 u}{\partial y \partial x^2} + \frac{\partial^3 u}{\partial y^3} \right) \\
&- \left( -\frac{1}{\rho} \cdot \frac{\partial^2 p}{\partial x \partial y} + \nu \left( \frac{\partial^3 v}{\partial x^3} + \frac{\partial^3 v}{\partial x \partial y^2} \right) \right)
\end{aligned}
\end{equation}

Considering the equality of mixed partial derivatives:

\[
\frac{\partial^2u}{\partial y\partial t}=\frac{\partial^2u}{\partial t\partial y},\quad \frac{\partial^2v}{\partial x\partial t}=\frac{\partial^2v}{\partial t\partial x},\quad \frac{\partial^2p}{\partial y\partial x}=\frac{\partial^2p}{\partial x\partial y}
\]

Some terms on the left-hand side of the equation can be simplified, and the pressure terms on the right-hand side cancel each other out. Using the definition of vorticity:

\[
\omega=\frac{\partial v}{\partial x}-\frac{\partial u}{\partial y}
\]

The subtracted equation can be further rearranged into the vorticity transport equation:

\[
\frac{\partial\omega}{\partial t}+u\frac{\partial\omega}{\partial x}+v\frac{\partial\omega}{\partial y}=\nu\left(\frac{\partial^2\omega}{\partial x^2}+\frac{\partial^2\omega}{\partial y^2}\right)
\]

This equation describes the rate of change of vorticity \(\omega\) with time \(t\), the convective terms of vorticity in space with fluid motion (\(u\frac{\partial\omega}{\partial x}+v\frac{\partial\omega}{\partial y}\)), and the diffusion term of vorticity (\(\nu\nabla^2\omega\)). It reflects the transport and diffusion processes of vorticity in the fluid.

\subsubsection{ Construction of the Loss Function}\label{subsubsec2}

In the OverPINNs method, the loss function is composed of several parts, including the residual loss of the original equations, the residual loss of the vorticity transport equation, and the losses of the initial and boundary conditions.

Let the predicted outputs of the neural network be \(u_\theta(x,y,t)\) and \(v_\theta(x,y,t)\), where \(\theta\) represents the parameters of the network. The total loss function \(L\) is defined as:

\[
L_{\mathrm{Total}}=L_{\mathrm{OE}}+L_{\mathrm{HE}}+L_{\mathrm{IC}}+L_{\mathrm{BC}}
\]

where:

- \(L_{\mathrm{OE}}\) is the loss of the original equations, i.e., the residual loss of the Navier-Stokes equations, calculated as:

\[
\begin{aligned}
L_{\mathrm{OE}} = \frac{1}{N_{\mathrm{colloc}}} \sum_{i=1}^{N_{\mathrm{colloc}}} \Biggl( \, & \left| \frac{\partial u}{\partial t} + u \frac{\partial u}{\partial x} + v \frac{\partial u}{\partial y} + \frac{1}{\rho} \frac{\partial p}{\partial x} - \nu \left( \frac{\partial^2 u}{\partial x^2} + \frac{\partial^2 u}{\partial y^2} \right) \right|_i^2 \\
& + \left| \frac{\partial v}{\partial t} + u \frac{\partial v}{\partial x} + v \frac{\partial v}{\partial y} + \frac{1}{\rho} \frac{\partial p}{\partial y} - \nu \left( \frac{\partial^2 v}{\partial x^2} + \frac{\partial^2 v}{\partial y^2} \right) \right|_i^2 \Biggr)
\end{aligned}
\]

- \(L_{\mathrm{HE}}\) is the loss of the vorticity transport equation, i.e., the residual loss of the newly derived vorticity transport equation, calculated as:

\[
L_{\mathrm{HE}}=\frac{1}{N_{\mathrm{colloc}}}\sum_{i=1}^{N_{\mathrm{colloc}}}\left(\left|\frac{\partial\omega}{\partial t}+u\frac{\partial\omega}{\partial x}+v\frac{\partial\omega}{\partial y}-\nu\left(\frac{\partial^2\omega}{\partial x^2}+\frac{\partial^2\omega}{\partial y^2}\right)\right|_i^2\right)
\]

- \(L_{\mathrm{IC}}\) is the loss of the initial conditions, calculated as:

\[
L_{\mathrm{IC}}=\frac{1}{N_{\mathrm{init}}}\sum_{i=1}^{N_{\mathrm{init}}}\left(\left|u\left(x_i^{\mathrm{init}},y_i^{\mathrm{init}},0\right)-u_{\mathrm{init}}\left(x_i^{\mathrm{init}},y_i^{\mathrm{init}}\right)\right|^2+\left|v\left(x_i^{\mathrm{init}},y_i^{\mathrm{init}},0\right)-v_{\mathrm{init}}\left(x_i^{\mathrm{init}},y_i^{\mathrm{init}}\right)\right|^2\right)
\]

- \(L_{\mathrm{BC}}\) is the loss of the boundary conditions, calculated as:

\[
\begin{aligned}
L_{\mathrm{BC}} = \frac{1}{N_{\mathrm{bound}}} \sum_{i=1}^{N_{\mathrm{bound}}} \Biggl( \, & \left| u\left(t_i^{\mathrm{bound}}, x_i^{\mathrm{bound}}, y_i^{\mathrm{bound}}\right) - u_{\mathrm{bound}}\left(t_i^{\mathrm{bound}}, x_i^{\mathrm{bound}}, y_i^{\mathrm{bound}}\right) \right|^2 \\
& + \left| v\left(t_i^{\mathrm{bound}}, x_i^{\mathrm{bound}}, y_i^{\mathrm{bound}}\right) - v_{\mathrm{bound}}\left(t_i^{\mathrm{bound}}, x_i^{\mathrm{bound}}, y_i^{\mathrm{bound}}\right) \right|^2 \Biggr)
\end{aligned}
\]

Here, \(N_{\mathrm{colloc}}\) is the number of points used to calculate the equation residuals, \(N_{\mathrm{init}}\) is the number of points used for the initial conditions, and \(N_{\mathrm{bound}}\) is the number of points used for the boundary conditions.

By imposing constraints from multiple aspects, OverPINNs can more accurately capture the physical laws embedded in the Navier-Stokes equations, thereby enhancing the accuracy and reliability of the solution.

\subsubsection{ Two-Dimensional Incompressible Vortex Evolution with Random Initial Velocity Field}\label{subsubsec2}

The incompressible fluid flow within a torus region typically exhibits complex vortex structures and periodic boundary conditions, which can effectively test the capability of PINNs to solve complex fluid dynamics problems. This problem can be described by the vorticity equation.

Vorticity transport equation:

\[
w_t+\mathbf{u}\cdot\nabla w=\frac{1}{Re}\Delta w,\ \mathrm{in\ }\left[0,T\right]\times\Omega
\]

This equation describes the evolution of vorticity \(w\) over time and space, where \(\mathbf{u}\) is the velocity field, \(Re\) is the Reynolds number representing the ratio of inertial forces to viscous forces in the fluid, and \(\Delta\) is the Laplacian operator.

Continuity equation:

\[
\nabla\cdot\mathbf{u}=0,\ \mathrm{in\ }\left[0,T\right]\times\Omega
\]

This equation ensures the incompressibility of the fluid, i.e., the divergence of the velocity field is zero.

Initial condition:

\[
w\left(0,x,y\right)=w_0\left(x,y\right),\mathrm{in\ }\Omega
\]

This provides the initial distribution of vorticity at time \(t=0\), giving the initial state of the problem.

To obtain the numerical solution, we implement periodic boundary conditions and an incompressible initial velocity field. The initial condition features a low-pass filtered random velocity distribution with constrained spectral content, as detailed in Table \ref{tab:ns_params}. Figure \ref{fig:mesh2} visualizes the corresponding vorticity field at \( t = 0 \).

\begin{table}[h]
\centering
\caption{Comprehensive fluid simulation parameters for Navier-Stokes validation}
\label{tab:ns_params}
\begin{tabular}{lcc}
\toprule
\textbf{Parameter} & \textbf{Value} & \textbf{Physical significance} \\
\midrule
Viscosity ($\nu$) & $1 \times 10^{-2}$ & Determines fluid viscous forces \\
Maximum velocity ($U_{\max}$) & 3 m/s & Upper bound for initial velocity field \\
Wavenumber cutoff & 2 & Spectral content limitation \\
Reynolds number ($Re$) & 100 & Inertial/viscous force ratio \\
Simulation duration & 10 s & Total evolution time \\
Computational grid & $1024\times1024$ & Spectral resolution covering $(0, 2\pi)^2$ \\
Stored resolution & $128\times128$ & Downsampled spatial resolution (8:1) \\
Temporal frames & 201 & Output time instances \\
\bottomrule
\end{tabular}
\end{table}

\begin{figure}[h]
    \centering
    \includegraphics[width=0.45\textwidth]{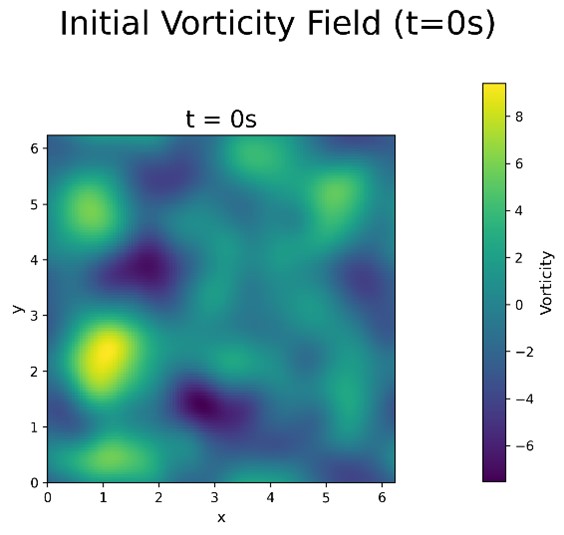}
    \caption{Initial vorticity distribution ($t = 0$ s)}
    \label{fig:mesh2}
\end{figure}

The numerical solution employs a hybrid Crank-Nicolson/RK4 scheme with spectral discretization:
\begin{itemize}
    \item \textbf{Viscous terms}: Implicit Crank-Nicolson ($O(\Delta t^2)$)
    \item \textbf{Convective terms}: Explicit 4th-order Runge-Kutta
    \item \textbf{Spectral operations}: RFFT for vorticity advancement
    \item \textbf{Poisson solver}: Pseudospectral method
\end{itemize}

Key implementation aspects:
\begin{itemize}
    \item Batch parallelization via \texttt{vmap} for temporal steps
    \item Spatial downsampling from $1024\times1024$ to $128\times128$
    \item Temporal resolution maintained at 201 frames
\end{itemize}

This high-fidelity numerical solution serves as the ground truth for evaluating PINN-based vorticity evolution predictions.

The numerical results are as follows:

\begin{figure}[h]
    \centering
    \includegraphics[width=1\textwidth]{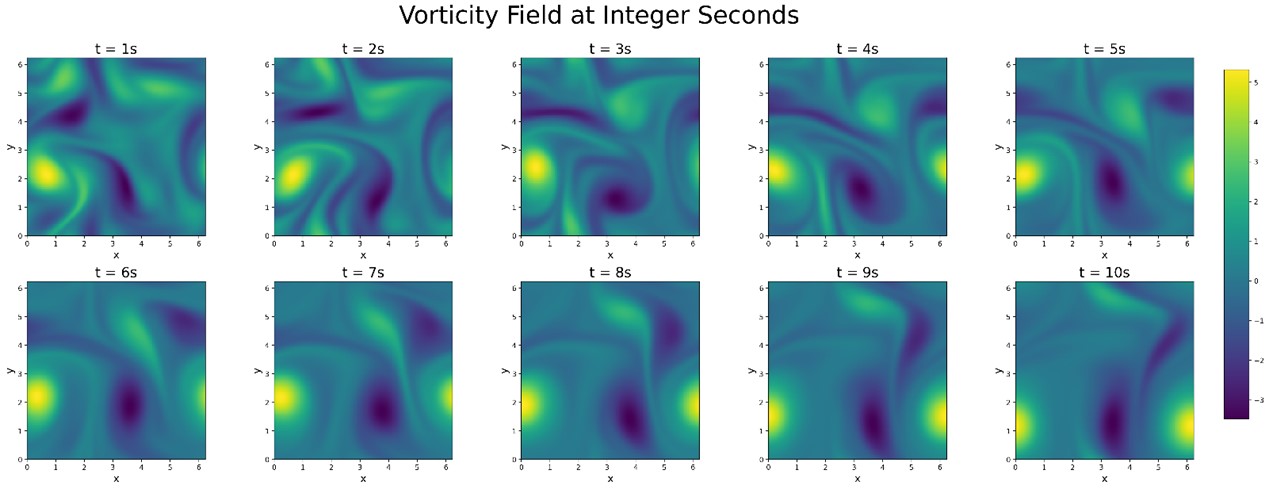}
    \caption{Evolution of the vorticity field at integer seconds (numerical solution).}
    \label{fig:mesh2}
\end{figure}

This section compares the performance of the original Physics-Informed Neural Network (PINN) and OverPINN in solving fluid dynamics problems. Both methods utilize a standard multilayer perceptron (MLP) architecture, with key structural parameters detailed in Table \ref{tab:params}. The network input dimension is 3, corresponding to the spatiotemporal coordinates \((t, x, y)\). The output dimension is either 2 or 3, depending on the formulation: PINNs using loss functions based solely on the Navier-Stokes (NS) equations predict velocity components \(u\) and \(v\) (output dim=2), while formulations incorporating the vorticity transport equation also predict vorticity \(\omega\) (output dim=3). Each network comprises 2 hidden layers with 256 neurons and uses the \(\tanh\) activation function. To enforce periodic boundary conditions, spatial coordinates are processed through a Fourier feature embedding layer with dimension 256. We evaluate the accuracy and reliability of both methods by analyzing errors in the velocity components \(u\), \(v\), and vorticity \(\omega\).

\begin{table}[h]
\centering
\caption{Neural Network Architecture Parameters}\label{tab:params}
\begin{tabular}{@{}lll@{}}
\toprule
\textbf{Parameter} & \textbf{Value} & \textbf{Description} \\
\midrule
Input Dimension & 3 & Spatiotemporal coordinates \((t, x, y)\) \\
Hidden Layer Dimension & 256 & Neurons per hidden layer \\
Number of Hidden Layers & 2 &  \\
Output Dimension & 2 \text{ or } 3 & Velocity components \((u, v)\) or \((u, v, \omega)\) \\
Activation Function & \(\tanh\) & Hidden layer activation \\
Network Architecture & MLP & Multi-layer perceptron \\
Fourier Embedding Dimension & 256 & Dimension of Fourier feature layer \\
\bottomrule
\end{tabular}
\footnotesize
\\[5pt] \textit{Source: Network parameters for the simulation model.}
\end{table}

For the original PINN, we conducted three comparative experiments with distinct loss function configurations:

1. Containing only the Navier-Stokes (NS) equations,

2. Containing only the vorticity transport equation,

3. Containing both the NS equations and the vorticity transport equation.

The corresponding total loss functions for the traditional PINN are:

NS Equations Only:
\[
L_{\mathrm{Total}} = L_{\mathrm{NS}} + L_{\mathrm{IC}} \tag{1}
\]

Vorticity Transport Equation Only:
\[
L_{\mathrm{Total}} = L_{\omega} + L_{\mathrm{IC}} \tag{2}
\]

\begin{figure}[htbp]
\centering
\includegraphics[width=0.65\textwidth]{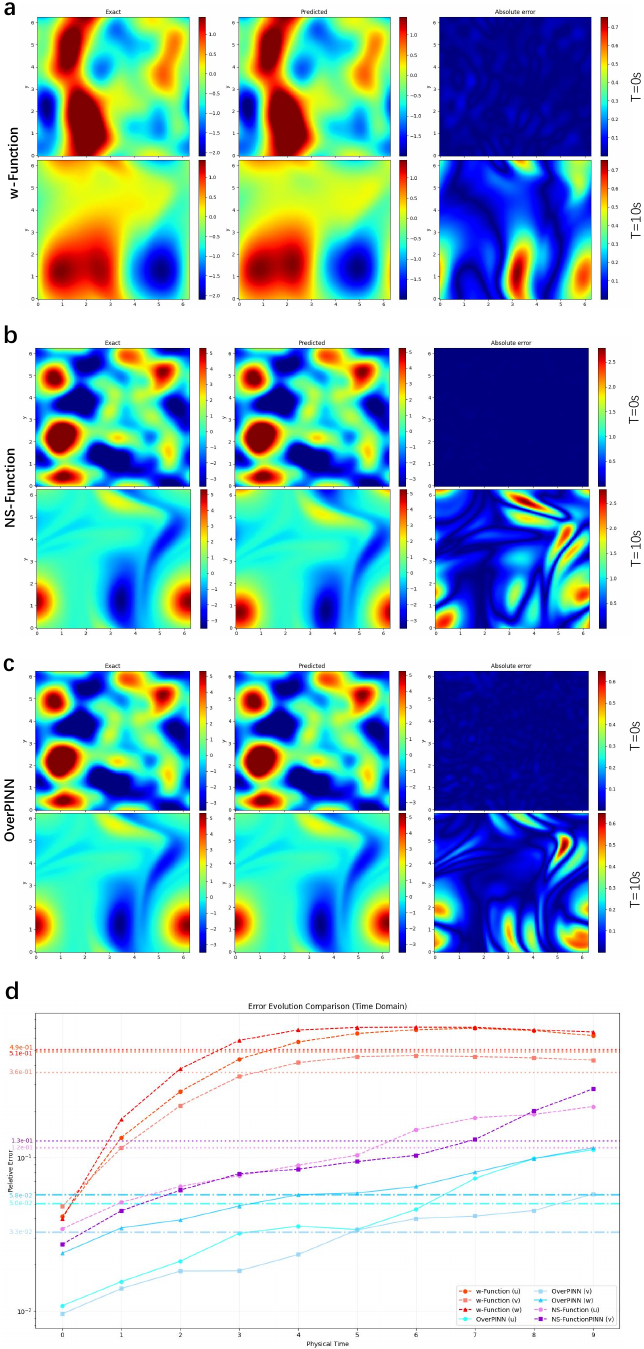}
\caption{Comparison of prediction results: (a) Traditional PINN using vorticity transport equation loss, showing exact vs. predicted velocity component $u$ and error distributions at $t = 0$ s and $t = 10$ s; (b) Traditional PINN using NS equations loss, showing vorticity $\omega$ predictions and errors; (c) OverPINN predictions for vorticity $\omega$; (d) Error evolution over time for all methods.}
\label{fig:results}
\end{figure}

The prediction results are presented in Fig.~\ref{fig:results}. 

\textbf{Key observations:}
\begin{itemize}
    \item \textbf{(a) Traditional PINN (vorticity-based loss):} Initial accurate fitting of velocity component $u$ deteriorates over time, showing significant errors at $t = 10$ s.
    
    \item \textbf{(b) Traditional PINN (NS equations loss):} Similar temporal error accumulation in vorticity $\omega$ predictions, reducing long-term accuracy.
    
    \item \textbf{(c) OverPINN:} Maintains high accuracy in vorticity $\omega$ predictions across both time points ($t = 0$ s and $t = 10$ s), with substantially reduced errors compared to traditional PINN.
    
    \item \textbf{(d) Error progression:} OverPINN demonstrates consistently lower errors and slower error growth over time, indicating enhanced stability and long-term predictive capability.
\end{itemize}

These results demonstrate that OverPINN significantly enhances solution accuracy by incorporating additional physical constraints. Beyond leveraging residuals from the governing equations, initial conditions, and boundary constraints, OverPINN systematically differentiates each partial differential equation within the system with respect to relevant variables. Through carefully designed algebraic manipulations, selected variables are eliminated to generate auxiliary equations. These supplementary physical constraints are incorporated into the composite loss function, enabling the model to simultaneously satisfy multi-faceted physical principles. This multi-constraint framework allows OverPINN to more accurately capture the underlying physics of fluid dynamics, particularly excelling in long-term transient simulations by effectively mitigating error propagation. The method thus provides an efficient and accurate approach for solving the Navier-Stokes equations, making it particularly suitable for high-precision fluid dynamics applications.

As a general framework, OverPINN demonstrates strong versatility in solving both individual PDEs and coupled PDE systems, achieving nearly an order-of-magnitude improvement in prediction accuracy for representative test cases. Crucially, the method maintains compatibility with established neural network optimization techniques without requiring complex equation derivations or specialized output modifications. This positions OverPINN as a generalizable solution for enhancing the accuracy of physics-informed neural networks across diverse partial differential equation systems.

\section{Conclusion}\label{sec2}

OverPINN achieves significant improvements in solving complex fluid dynamics problems, demonstrating considerable potential for enhancing spatiotemporal dynamics prediction accuracy. By incorporating supplementary physical constraints through high-order PDE formulations, OverPINN substantially enhances prediction fidelity compared to traditional PINN methods. This advancement holds particular value for applications requiring high-precision simulations, such as turbulence modeling and reaction-diffusion process analysis.

A key advantage of OverPINN lies in its ability to integrate multi-layered physical constraints into the learning framework. Unlike conventional PINNs that primarily rely on original governing equations, OverPINN systematically generates auxiliary equations through carefully designed differentiation and algebraic operations. For instance, in fluid dynamics, OverPINN derives the vorticity transport equation directly from the Navier-Stokes equations. This approach enriches the neural network's training information while providing a more comprehensive physical representation, enabling more accurate capture of complex phenomena including vorticity evolution and energy transfer.

OverPINN effectively addresses the persistent challenge of temporal error accumulation in evolving systems by constraining the solution space with supplementary physical laws. The auxiliary equations incorporated into the loss function serve as physics-informed regularizers, guiding the network toward solutions that better adhere to underlying physical principles. Consequently, OverPINN demonstrates superior stability and accuracy in long-term predictions, as evidenced in both Allen-Cahn equation simulations—where it accurately resolves rapid gradient changes and interface dynamics—and Navier-Stokes applications—where it significantly mitigates error propagation over extended durations.

The method preserves the implementation flexibility that has popularized PINN approaches, while maintaining compatibility with established neural network optimization techniques. This seamless integration enables performance enhancement without requiring architectural modifications or output restructuring, positioning OverPINN as a practical and versatile tool for scientific computing.

OverPINN opens new avenues for addressing multi-physics and multi-scale challenges. By systematically incorporating equations governing diverse physical processes and scales, the framework shows promise for delivering enhanced solutions in domains such as climate modeling, materials science, and biological systems. For example, in climate simulations, OverPINN could concurrently represent processes spanning multiple spatiotemporal scales—from macroscale atmospheric circulation to microscale cloud microphysics—thereby improving overall model predictive capability.

This framework advances understanding of physics-informed neural networks in scientific computing, highlighting the importance of structured domain knowledge integration and providing a methodological approach for enhancing model performance. Future research could explore more sophisticated strategies for embedding physical laws within neural architectures, further expanding the potential of physics-informed machine learning.

Despite its demonstrated efficacy for complex scientific problems, OverPINN implementation presents several challenges. A primary concern involves the judicious selection and formulation of supplementary physical constraints. While incorporating high-order derivatives of governing equations improves accuracy, these terms substantially increase computational complexity. The resulting sophisticated loss functions demand greater computational resources, potentially leading to prohibitive memory requirements—particularly for large-scale simulations or high-resolution models. These resource demands may constrain widespread adoption in environments lacking high-performance computing infrastructure. Balancing accuracy gains against computational constraints remains crucial for broader application across scientific domains. Addressing these computational challenges will be essential to fully realize OverPINN's potential across diverse research contexts.

\bibliographystyle{unsrt}  
\bibliography{references}

\begin{thebibliography}{10}

\bibitem{0Promising}
Steven~L. Brunton and J.~Nathan Kutz.
\newblock Promising directions of machine learning for partial differential equations.
\newblock {\em Nature Computational Science}.

\bibitem{pde2}
Machine learning solutions looking for pde problems.
\newblock {\em Nature Machine Intelligence}, 7(1):1--1, jan 01 2025.

\bibitem{Brandstetter2025}
Johannes Brandstetter.
\newblock Envisioning better benchmarks for machine learning pde solvers.
\newblock {\em Nature Machine Intelligence}, 7(1):2--3, jan 2025.

\bibitem{1992Numerical}
Randall~J. Leveque.
\newblock {\em Numerical Methods for Conservation Laws}.
\newblock Numerical Methods for Conservation Laws, 1992.

\bibitem{finitedifference}
Feng Guo and Luis Couto.
\newblock Comparative performance analysis of numerical discretization methods for electrochemical model of lithium-ion batteries.
\newblock {\em Journal of Power Sources}, 650:237365, 09 2025.

\bibitem{FDE1}
Mark~M. Meerschaert and Charles Tadjeran.
\newblock Finite difference approximations for fractional advection–dispersion flow equations.
\newblock {\em Journal of Computational and Applied Mathematics}, 172(1):65--77, 2004.

\bibitem{FDE2}
Anatoly~A. Alikhanov.
\newblock A new difference scheme for the time fractional diffusion equation.
\newblock {\em Journal of Computational Physics}, 280:424--438, 2015.

\bibitem{FDE3}
Guang-Hua Gao, Hai-Wei Sun, and Zhi-Zhong Sun.
\newblock Stability and convergence of finite difference schemes for a class of time-fractional sub-diffusion equations based on certain superconvergence.
\newblock {\em Journal of Computational Physics}, 280:510--528, 2015.

\bibitem{FDE4}
B.P. Moghaddam and J.A.T. Machado.
\newblock A stable three-level explicit spline finite difference scheme for a class of nonlinear time variable order fractional partial differential equations.
\newblock {\em Computers \& Mathematics with Applications}, 73(6):1262--1269, 2017.
\newblock Advances in Fractional Differential Equations (IV): Time-fractional PDEs.

\bibitem{FDE5}
Sekar Elango, Ayyadurai Tamilselvan, R.~Vadivel, Nallappan Gunasekaran, Haitao Zhu, Jinde Cao, and Xiaodi Li.
\newblock Finite difference scheme for singularly perturbed reaction diffusion problem of partial delay differential equation with nonlocal boundary condition.
\newblock {\em Advances in Difference Equations}, 2021(1):151, mar 04 2021.

\bibitem{FDE0}
Sergei~K. Godunov and I.~Bohachevsky.
\newblock Finite difference method for numerical computation of discontinuous solutions of the equations of fluid dynamics.
\newblock {\em Matematičeskij sbornik}, 47(89)(3):271--306, 1959.
\newblock \href{https://hal.archives-ouvertes.fr/hal-01620642}{hal-01620642}.

\bibitem{FEM}
Robert Anderson, Julian Andrej, Andrew Barker, Jamie Bramwell, Jean-Sylvain Camier, Jakub Cerveny, Veselin Dobrev, Yohann Dudouit, Aaron Fisher, Tzanio Kolev, Will Pazner, Mark Stowell, Vladimir Tomov, Ido Akkerman, Johann Dahm, David Medina, and Stefano Zampini.
\newblock Mfem: A modular finite element methods library.
\newblock {\em COMPUTERS \& MATHEMATICS WITH APPLICATIONS}, 81(SI):42--74, JAN 1 2021.

\bibitem{FEM1}
Yingjun Jiang and Jingtang Ma.
\newblock High-order finite element methods for time-fractional partial differential equations.
\newblock {\em Journal of Computational and Applied Mathematics}, 235(11):3285--3290, 2011.

\bibitem{FEM2}
Max~D. Gunzburger, Clayton~G. Webster, and Guannan Zhang.
\newblock Stochastic finite element methods for partial differential equations with random input data.
\newblock {\em Acta Numerica}, 23:521–650, 2014.

\bibitem{FEM3}
Changpin Li and Zhen Wang.
\newblock The local discontinuous galerkin finite element methods for caputo-type partial differential equations: Mathematical analysis.
\newblock {\em Applied Numerical Mathematics}, 150:587--606, 2020.

\bibitem{FEM0}
O.~C. Zienkiewicz, R.~L. Taylor, P.~Nithiarasu, and J.~Z. Zhu.
\newblock {\em The Finite Element Method}, volume~3.
\newblock Elsevier, 1977.

\bibitem{fv}
Robert Eymard, Thierry Gallou\"et, and Rapha\`ele Herbin.
\newblock Finite volume methods.
\newblock In {\em Handbook of numerical analysis, {V}ol. {VII}}, volume VII of {\em Handb. Numer. Anal.}, pages 713--1020. North-Holland, Amsterdam, 2000.

\bibitem{spectral-methods}
Waleed~Mohamed Abd-Elhameed and Hany~Mostafa Ahmed.
\newblock Spectral solutions for the time-fractional heat differential equation through a novel unified sequence of chebyshev polynomials.
\newblock {\em AIMS MATHEMATICS}, 9(1):2137--2166, 2024.

\bibitem{MR678711}
M.~D. Feit, J.~A. Fleck, Jr., and A.~Steiger.
\newblock Solution of the {S}chr\"odinger equation by a spectral method.
\newblock {\em J. Comput. Phys.}, 47(3):412--433, 1982.

\bibitem{Spectral}
John Boyd.
\newblock Spectral methods in fluid dynamics (c. canuto, m. y. hussaini, a. quarteroni, and t. a. zang).
\newblock {\em Siam Review - SIAM REV}, 30, 12 1988.

\bibitem{2013Sparse}
Hayden Schaeffer, Russel Caflisch, Cory~D. Hauck, and Stanley Osher.
\newblock Sparse dynamics for partial differential equations.
\newblock {\em Proc Natl Acad Sci U S A}, 110(17):6634--6639, 2013.

\bibitem{2021Highly}
John Jumper, Richard Evans, Alexander Pritzel, Tim Green, and Demis Hassabis.
\newblock Highly accurate protein structure prediction with alphafold.
\newblock {\em Nature}, pages 1--11, 2021.

\bibitem{2021Deep}
Pranshu Pant, Ruchit Doshi, Pranav Bahl, and Amir~Barati Farimani.
\newblock Deep learning for reduced order modelling and efficient temporal evolution of fluid simulations.
\newblock 2021.

\bibitem{2021Physics}
Xiaowei Jia, Jared Willard, Anuj Karpatne, Jordan~S. Read, Jacob~A. Zwart, Michael Steinbach, and Vipin Kumar.
\newblock Physics-guided machine learning for scientific discovery: An application in simulating lake temperature profiles.
\newblock {\em ACM/IMS Transactions on Data Science}, (3), 2021.

\bibitem{2020Integrating}
Jared Willard, Xiaowei Jia, Shaoming Xu, Michael Steinbach, and Vipin Kumar.
\newblock Integrating scientific knowledge with machine learning for engineering and environmental systems.
\newblock 2020.

\bibitem{2017Meaningless}
De~Masi Orianna, Kording Konrad, Recht Benjamin, and Jan Yih-Kuen.
\newblock Meaningless comparisons lead to false optimism in medical machine learning.
\newblock {\em Plos One}, 12(9), 2017.

\bibitem{2024Weak}
Nick Mcgreivy and Ammar Hakim.
\newblock Weak baselines and reporting biases lead to overoptimism in machine learning for fluid-related partial differential equations.
\newblock 2024.

\bibitem{2016Best}
B.~Wujek and Patrick Hall.
\newblock Best practices for machine learning applications.
\newblock 2016.

\bibitem{RAISSI2019686}
M.~Raissi, P.~Perdikaris, and G.E. Karniadakis.
\newblock Physics-informed neural networks: A deep learning framework for solving forward and inverse problems involving nonlinear partial differential equations.
\newblock {\em Journal of Computational Physics}, 378:686--707, 2019.

\bibitem{Rao2023}
Chengping Rao, Pu~Ren, Qi~Wang, Oral Buyukozturk, Hao Sun, and Yang Liu.
\newblock Encoding physics to learn reaction–diffusion processes.
\newblock {\em Nature Machine Intelligence}, 5(7):765--779, jul 2023.

\bibitem{KHARAZMI2021113547}
Ehsan Kharazmi, Zhongqiang Zhang, and George~E.M. Karniadakis.
\newblock hp-vpinns: Variational physics-informed neural networks with domain decomposition.
\newblock {\em Computer Methods in Applied Mechanics and Engineering}, 374:113547, 2021.

\bibitem{Jahani-Nasab2024}
Jahani-Nasab M and Bijarchi MA.
\newblock Enhancing convergence speed with feature enforcing physics-informed neural networks using boundary conditions as prior knowledge.
\newblock {\em Scientific Reports}, 14(1):23836, oct 11 2024.

\bibitem{MCCLENNY2023111722}
Levi~D. McClenny and Ulisses~M. Braga-Neto.
\newblock Self-adaptive physics-informed neural networks.
\newblock {\em Journal of Computational Physics}, 474:111722, 2023.

\bibitem{CiCP-29-3}
Colby Wight, L. and Jia Zhao.
\newblock Solving allen-cahn and cahn-hilliard equations using the adaptive physics informed neural networks.
\newblock {\em Communications in Computational Physics}, 29(3):930--954, 2021.

\bibitem{timewindows}
Cheng Su, Jingwei Liang, and Zengsheng He.
\newblock E-pinn: A fast physics-informed neural network based on explicit time-domain method for dynamic response prediction of nonlinear structures.
\newblock {\em Engineering Structures}, 321:118900, 2024.

\bibitem{tw1}
Fangyu Liu, Junlin Li, and Linbing Wang.
\newblock Pi-lstm: Physics-informed long short-term memory network for structural response modeling.
\newblock {\em Engineering Structures}, 292:116500, 2023.

\bibitem{tw2}
Ruiyang Zhang, Yang Liu, and Hao Sun.
\newblock Physics-informed multi-lstm networks for metamodeling of nonlinear structures.
\newblock {\em Computer Methods in Applied Mechanics and Engineering}, 369:113226, 2020.

\bibitem{tw3}
Xuhui Meng, Zhen Li, Dongkun Zhang, and George~Em Karniadakis.
\newblock Ppinn: Parareal physics-informed neural network for time-dependent pdes.
\newblock {\em Computer Methods in Applied Mechanics and Engineering}, 370:113250, 2020.

\bibitem{tw5}
Ameya Jagtap, D. and Em~Karniadakis, George.
\newblock Extended physics-informed neural networks (xpinns): A generalized space-time domain decomposition based deep learning framework for nonlinear partial differential equations.
\newblock {\em Communications in Computational Physics}, 28(5):2002--2041, 2020.

\bibitem{2020Adaptive}
Ameya~D. Jagtap, Kenji Kawaguchi, and George~Em Karniadakis.
\newblock Adaptive activation functions accelerate convergence in deep and physics-informed neural networks.
\newblock {\em Journal of Computational Physics}, 2020.

\bibitem{2021Deep1}
Chen Cheng and Guang~Tao Zhang.
\newblock Deep learning method based on physics informed neural network with resnet block for solving fluid flow problems.
\newblock {\em Water}, (4), 2021.

\bibitem{wang2020understanding}
Sifan Wang, Yujun Teng, and Paris Perdikaris.
\newblock Understanding and mitigating gradient pathologies in physics-informed neural networks.
\newblock {\em arXiv preprint arXiv:2001.04536}, 2020.

\bibitem{hard}
Liu Ziqi, Cai Wei, and John Zhi-Qin, Xu.
\newblock Multi-scale deep neural network (mscalednn) for solving poisson-boltzmann equation in complex domains.
\newblock {\em Communications in Computational Physics}, 28(5):1970--2001, 2020.

\bibitem{ad}
Atilim Baydin, Barak Pearlmutter, Alexey Radul, and Jeffrey Siskind.
\newblock Automatic differentiation in machine learning: A survey.
\newblock {\em Journal of Machine Learning Research}, 18:1--43, 04 2018.

\bibitem{doi:10.1137/19M1274067}
Lu~Lu, Xuhui Meng, Zhiping Mao, and George~Em Karniadakis.
\newblock Deepxde: A deep learning library for solving differential equations.
\newblock {\em SIAM Review}, 63(1):208--228, 2021.

\bibitem{Buescu2021}
Jorge Buescu and Cristina Serpa.
\newblock Compatibility conditions for systems of iterative functional equations with non-trivial contact sets.
\newblock {\em Results in Mathematics}, 76(2):68, mar 17 2021.

\end{thebibliography}

\end{document}